\documentclass[conference]{IEEEtran}
\IEEEoverridecommandlockouts
\usepackage{cite}
\usepackage{amsmath,amssymb,amsfonts}
\usepackage{algorithm}
\usepackage{algpseudocode}
\usepackage{booktabs,multirow,multicol}

\usepackage{graphicx}
\usepackage{textcomp}
\usepackage{xcolor}
\usepackage{soul}
\usepackage{scalerel}
\usepackage{stackengine}
\usepackage{multirow}

\usepackage{color}

\def\BibTeX{{\rm B\kern-.05em{\sc i\kern-.025em b}\kern-.08em
    T\kern-.1667em\lower.7ex\hbox{E}\kern-.125emX}}

\usepackage[colorlinks=true,linkcolor=red,citecolor=blue, urlcolor=blue]{hyperref}
\usepackage{tabularray}

\newlength{\blob}
\title{Unsupervised Feature Clustering Improves Contrastive Representation Learning for \\Medical Image Segmentation
\thanks{* indicates equal contribution.\\
\indent 978-1-6654-6819-0/22/\$31.00 ©2022 IEEE
}
}
\makeatletter
\newcommand{\linebreakand}{%
  \end{@IEEEauthorhalign}
  \vspace{-3mm}
  \hfill\mbox{}\par
  \mbox{}\hfill\begin{@IEEEauthorhalign}
}
\makeatother

\author{
\IEEEauthorblockN{Yejia Zhang*}
\IEEEauthorblockA{\textit{University of Notre Dame} \\
                  Notre Dame, IN, 46556, USA \\
                  yzhang46@nd.edu} 
\and
\IEEEauthorblockN{Xinrong Hu*}
\IEEEauthorblockA{\textit{University of Notre Dame} \\
                  Notre Dame, IN, 46556, USA \\
                  xhu7@nd.edu} 
\and
\IEEEauthorblockN{Nishchal Sapkota}
\IEEEauthorblockA{\textit{University of Notre Dame} \\
                  Notre Dame, IN, 46556, USA \\
                  nsapkota@nd.edu}
\linebreakand 
\IEEEauthorblockN{Yiyu Shi}
\IEEEauthorblockA{\textit{University of Notre Dame} \\
                  Notre Dame, IN, 46556, USA \\
                  yshi4@nd.edu}
\and
\IEEEauthorblockN{Danny Z. Chen}
\IEEEauthorblockA{\textit{University of Notre Dame} \\
                  Notre Dame, IN, 46556, USA \\
                  dchen@nd.edu}
}

\makeatletter
\patchcmd{\@maketitle}
  {\addvspace{0.5\baselineskip}\egroup}
  {\addvspace{-1.2\baselineskip}\egroup}
  {}
  {}
\makeatother

\begin{document}

\maketitle

\begin{abstract}

Self-supervised instance discrimination is an effective contrastive pretext task to learn feature representations and address limited medical image annotations.
The idea is to make features of transformed versions of the same images similar while forcing all other augmented images' representations to contrast. 
However, this instance-based contrastive learning leaves performance on the table by failing to maximize feature affinity between images with similar content while counter-productively pushing their representations apart. 
Recent improvements on this paradigm (e.g., leveraging multi-modal data, different images in longitudinal studies, spatial correspondences) either relied on additional views or made stringent assumptions about data properties, which can sacrifice generalizability and applicability. 
To address this challenge, we propose a 
new self-supervised contrastive learning method that uses unsupervised feature clustering to better select positive and negative image samples.
More specifically, we produce pseudo-classes by hierarchically clustering features obtained by an auto-encoder in an unsupervised manner, and prevent destructive interference during contrastive learning by avoiding the selection of negatives from the same pseudo-class.
Experiments on 2D skin dermoscopic image segmentation and 3D multi-class whole heart CT segmentation demonstrate 
that our method outperforms state-of-the-art self-supervised contrastive techniques on these tasks.

\end{abstract}


\section{Introduction}
\label{sec:introdcution}

\begin{figure}[t]
    \centering
    \includegraphics[width=1.0\linewidth]{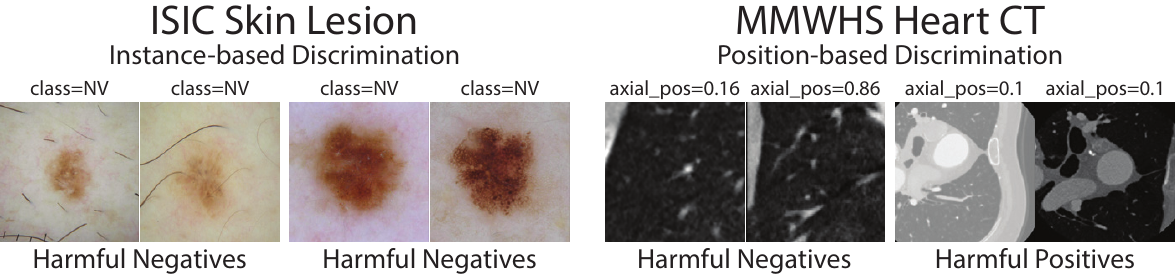}
    \vspace*{-5.5mm}
    \caption{Examples of harmful selections of positives and negatives from our downstream datasets. 
    On the \textbf{left}, 
    skin lesions with similar appearances and the same class label are treated as negatives (harmful negatives). 
    On the \textbf{right}, choosing positive and negative slices based on axial position proximity leads to both harmful negatives (similar lung tissue in distant CT slices) and harmful positives (proximal slices across two volumes still vary greatly).}
    \label{fig:motivation}
\vspace*{-5.5mm}
\end{figure}

Self-supervised contrastive methods have recently made great strides in representation learning both in natural scene imaging \cite{chen2020simple,chen2020mocov2} and biomedical imaging \cite{Xie2020PGLPL,Azizi2021BigSM,Sowrirajan2021MoCoPI,Zheng2021HierarchicalSL,Chen2021MomentumCL} applications.
The core idea is to pull representations of ``similar" images (referred to as \emph{positives}) close while ``dissimilar" images (\emph{negatives}) are contrasted in feature space. 
Such methods implemented this idea using an instance discrimination pretext task where only transformed versions of the same images are taken as positives while augmented versions of other images are negatives.
This approach is sub-optimal when two different images that contain similar content are 
treated as negatives (e.g., see ``Harmful Negatives" in Fig.~\ref{fig:motivation}).
This disproportionately hinders medical imaging pretraining compared to natural scene datasets since medical image datasets generally have far fewer semantic categories, less inter-class as well as intra-class appearance diversity, and considerably fewer samples to learn meaningful features.

Recent \textbf{contrastive learning methods for biomedical imaging} tasks addressed these drawbacks by improving the selection of positives and negatives.
One common approach is to designate different images of the same patient as positives.
This can be done either using images of a patient across time in longitudinal studies \cite{Jamaludin2017SelfsupervisedLF,Sriram2021COVID19PV} or leveraging scans taken at the same time but from different views \cite{Azizi2021BigSM,Vu2021MedAugCL}.
A related idea is to treat the same anatomical structure captured from different modalities as positives \cite{li2020self}.
However, these methods require the existence of relevant meta-data and additional corresponding images which are often unavailable.
Instead of relying on instance correspondences, Global \& Local Contrastive Learning (GLCL \cite{chaitanya2020contrastive}) and Positional Contrastive Learning (PCL \cite{zeng2021positional}) proposed to use slices in close proximity as positives by assuming semantic correlations in corresponding axial positions of 3D volumes.
These approaches reasonably avoid some harmful negatives (although many remain unaddressed; see MMWHS ``Harmful Negatives" in Fig.~\ref{fig:motivation}) and benefit from sample diversity in positives.
However, they assumed that images are sensibly aligned (see ``Harmful Positives" in Fig.~\ref{fig:motivation} for an example of poor alignment) and, although they operate in 2D slices, cannot be applied to general 2D tasks.
These advances show promising potential of contrastive learning beyond instance discrimination, but they either have stringent data correspondence requirements or rely on binding assumptions of image structures which limit their generalizability and 
applicability to other biomedical tasks.
Furthermore, many rely on heuristics that still do not fully handle the issue of harmful positives \& negatives.

To address these challenges, we explore distances between learned image features as a proxy for appearance similarity.
Utilizing feature distances enables us to cluster images based on appearance in a way that is effective across 
tasks, imaging modalities, data dimensionality, and spatial settings.
In \textbf{related medical imaging works}, 
\cite{zheng2019biomedical} used variational auto-encoders to attain features of images to select representative samples for improving annotation efficiency, and
\cite{Haghighi2020LearningSR} used features from an auto-encoder to select similar instances to assign corresponding anatomical regions for pretraining. 
Our objective is to leverage the extracted pseudo-categories in order to improve samplings of positives \& negatives for contrastive learning.
We choose to use Beta Variational Auto-Encoders ($\beta$-VAEs) \cite{higgins2016beta} for their general applicability to image data, ease of training, and propensity to learn disentangled features.

In this paper, we propose a contrastive learning framework based on unsupervised feature clustering (called \textbf{UFC}), which outperforms instance discrimination pretraining and can be employed generally across medical imaging modalities \& tasks. 
First, we extract categorical information from unlabeled data in an unsupervised manner by hierarchically clustering features learned by a $\beta$-VAE.
Second, we pretrain a convolutional encoder with contrastive learning guided by pseudo-categories from the predicted clusters to avoid contrasting similar images while maximizing similarity with more diverse positive samples.
Finally, we transfer the encoder to fine-tune on a downstream segmentation task to assess the quality of features learned. 
We evaluate our framework on
a skin lesion 
dataset (ISIC) and a heart CT 
dataset (MMWHS) for segmentation. 
The results show that our framework outperforms state-of-the-art contrastive learning methods irrespective of the percentages of annotations used for fine-tuning.

\begin{figure}
    \centering
    \includegraphics[width=0.484\textwidth]{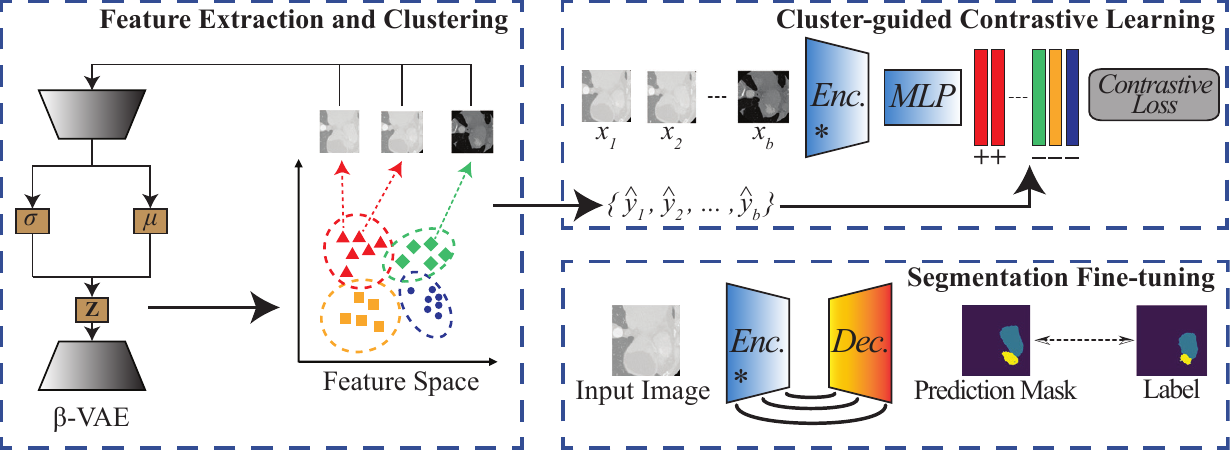}   
    \vspace*{-7mm}
    \caption{\small{The overall workflow of our proposed framework. 
    }}
    \label{fig:framework}
    \vspace*{-5.5mm}
\end{figure}

\begin{table*}
\begin{center}
\caption{\label{table1}Comparisons against state-of-the-art contrastive methods in Dice score {\%}'s {\small (STD)} using different amounts of training annotations, averaged over 5 runs.
LGCL \& PCL are designed for 3D volumes and are not applicable to ISIC. 
}
\vspace*{-2.5mm}
\resizebox{0.8\textwidth}{!}{
\begin{tabular}{c||c c c c | c c c c}
\hline 
 \rule[0pt]{0pt}{\heightof{A}+1ex}\multirow{2}{*} {Method}
& \multicolumn{4}{c} {ISIC Skin Dermoscopy} & 
\multicolumn{4}{c} {MMWHS Heart CT} \\ [0.3em]
\cline{2-9} 
 \rule[0pt]{0pt}{\heightof{A}+1ex}& 5\% & 10\% & 20\% & 100\% & 10\% & 20\% & 40\% &100\% \\ [0.3em]
 \hline
  \rule[0pt]{0pt}{\heightof{A}+1ex}Rand. Init. & 79.3 ($\pm$ 0.3) & 81.7 ($\pm$ 0.2) & 83.3 ($\pm$ 0.2) & 84.7 ($\pm$ 0.5) & 38.4 ($\pm$ 6.2) & 62.1 ($\pm$ 7.0) & 73.8 ($\pm$ 4.9) & 85.9 ($\pm$ 2.7)  \\[0.3em]

 SimCLR \cite{chen2020simple}  & 80.1 ($\pm$ 0.8) & 82.7 ($\pm$ 0.3) & 83.7 ($\pm$ 0.2) & 84.9 ($\pm$ 0.3) & 44.5 ($\pm$ 8.4) & 67.2 ($\pm$ 8.3) & 80.6 ($\pm$ 2.9) & 86.6 ($\pm$ 2.5)  \\[0.3em]
 
 MoCov2 \cite{chen2020mocov2}  & 79.9 ($\pm$ 0.7) & 82.2 ($\pm$ 0.7) & 83.3 ($\pm$ 0.2) & 84.8 ($\pm$ 0.2) & 42.4 ($\pm$ 7.2) & 64.7 ($\pm$ 9.7) & 75.9 ($\pm$ 5.3) & 86.4 ($\pm$ 2.3)  \\[0.3em]
 
 BYOL \cite{grill2020bootstrap} & 79.4 ($\pm$ 0.5) & 80.6 ($\pm$ 0.4) & 82.5 ($\pm$ 0.4) & 84.8 ($\pm$ 0.3) & 42.0 ($\pm$ 5.0) & 58.5 ($\pm$ 10.7) & 69.5 ($\pm$ 9.1) & 86.1 ($\pm$ 2.0)  \\[0.3em]
 
 \hline 
  \rule[0pt]{0pt}{\heightof{A}+1ex}LGCL \cite{chaitanya2020contrastive}  & -  & - & - & - & 35.5 ($\pm$ 2.5) & 62.4 ($\pm$ 12.1) & 74.0 ($\pm$ 3.8) & 85.2 ($\pm$ 2.4)  \\[0.3em]
 
 PCL \cite{zeng2021positional}  & -  & - & - & - & 38.2 ($\pm$ 6.5) & 60.8 ($\pm$ 8.3) & 71.6 ($\pm$ 9.2) & 85.5 ($\pm$ 2.8)  \\[0.3em]
 
 \hline
   \rule[0pt]{0pt}{\heightof{A}+1ex}Ours & \textbf{81.9 ($\pm$ 0.4)} & \textbf{82.8 ($\pm$ 0.3)} & \textbf{84.6 ($\pm$ 0.4)} & \textbf{85.3 ($\pm$ 0.2)} & \textbf{46.9 ($\pm$ 5.8)} & \textbf{69.9 ($\pm$ 8.2)} & \textbf{81.4 ($\pm$ 2.6)} & \textbf{87.3 ($\pm$ 1.5)}  \\[0.3em]
   \hline
\end{tabular}
}
\end{center}
\vspace*{-7.5mm}
\end{table*}

Our main \textbf{contributions} can be summarized below.
\begin{enumerate}
    \item We propose a novel three-stage self-supervised contrastive learning framework that is broadly applicable, effective for diverse medical image data, and easily incorporated into existing deep learning pipelines.
    \item Given the limited works that address the difficulties of harmful negatives in medical imaging tasks, we show the potential of using categorical information extracted for ``free" with only unlabeled data to improve representation learning for medical image segmentation, and highlight the disadvantages of instance discrimination.
    \item We achieve state-of-the-art performances on the skin lesion segmentation and multi-class CT heart segmentation tasks 
    over known contrastive methods.
\end{enumerate}

\section{Method}
\label{sec:methods}

Our method consists of three main stages: unsupervised feature clustering (UFC), cluster-guided contrastive learning, and segmentation fine-tuning (see Fig.~\ref{fig:framework}). 
First, we train a $\beta$-VAE to learn feature representations in an unsupervised manner and apply a hierarchical clustering method to divide those image-level features 
into clusters.
Next, we assign features in each cluster the same pseudo-label, which guides the contrastive learning of our encoder $Enc$.
The above two stages serve as the pretraining process. 
Finally, we keep the weights in $Enc$ and concatenate it with a decoder $Dec$ to fine-tune the downstream segmentation task.
We detail 
UFC and cluster-guided contrast learning in the following subsections.

\subsection{Unsupervised Feature Clustering} \label{sec:methods1}
To group images that share similar anatomical structures more easily, the first step is to map the images to a low-dimensional feature space.
Many unsupervised methods can be utilized, including a pretext task, contrastive learning (CL), an auto-encoder (AE), and a variational auto-encoder (VAE). 
Regarding the pretext task, training the model to predict the rotation or positions of input images results in extracting task-relevant features with essential information missed. 
For CL, previous works like \cite{chaitanya2020contrastive} demonstrated reasonable initializations for segmentation tasks, but the biases incurred from harmful samples could magnify if we use their features directly for clustering.
Hence, we mainly focus on the AE \& VAE, which both comprise of an encoder $E(\cdot)$ and a decoder $D(\cdot)$.
For an input image $m_{i}$, the encoder gives a feature representation of $m_{i}$, $z_{i} = E(m_{i})$, and the decoder further outputs a reconstructed image, $\hat{m}_{i} = D(E(m_{i}))$. 
AE learns the embeddings only by minimizing the mean square error between $m_{i}$ and $\hat{m}_{i}$, while VAE assumes the distribution of the embeddings to be a Gaussian distribution $p(z) = G(\mu, \theta)$ and seeks to force the real distribution of $z$ to be close to this posterior distribution. 
More specifically, the encoder of VAE generates both a mean vector $\hat{\mu}$ and a variance vector $\hat{\theta}$. After that, the feature vector $z$ is sampled from the distribution $q(z|x) = G(\hat{\mu}, \hat{\theta})$ for image reconstruction. 
The overall loss function for VAE is:
\begin{equation} \label{vae_loss}
\begin{split}
    L_{vae} & = L_{rec} + L_{prior} \\
            & = \mathbb{E}_{q(z|x)}[\log(p(x|z)] - \beta * D_{KL}(q(z|x) \| p(z)),
\end{split}
\end{equation}
where $D_{KL}$ is the Kullback-Leibler divergence and
$\beta$ is a weight to adjust how strong the constraint should be put on the latent distribution \cite{higgins2016beta}. If $\beta = 1$, then the loss in Eq.~(\ref{vae_loss}) corresponds to the original VAE; otherwise, it is $\beta$-VAE.

After features are obtained, clustering them into appearance-based pseudo-categories is a key step in UFC. 
We use the agglomerative clustering method \cite{zheng2019biomedical} that hierarchically combines features in the closest proximity so that images that look the most similar would be grouped first.
We avoid centroid-based clustering since they have an inherent disadvantage of having to select the number of centers which is very task-dependent. 
This issue is mitigated in density-based clustering; however, we must assume that data form coherent density patterns in embedding space which may not be the case given imbalanced categories with limited examples for classes of interest or data distributions with large intra-class appearance variations.
Furthermore, for distribution-based techniques, prior knowledge of the distribution is needed, which is difficult in high-dimensional objects like images.
Thus, we find agglomerative clustering to be well-suited for our task given its flexibility in selecting cluster sizes as well as its tendency to prioritize the most visually-similar images for grouping.

In order to decide the number of clusters, $k$, we draw a curve of mean distance within each cluster vs.~$k$, and the elbow point roughly gives a range for a $k$ candidate. 
Actually, we find that the value of $k$ largely depends on the dataset, and there is no optimal value, rather there is just an optimal range for most cases.
We assign every image embedding to a cluster in the feature space. Based on that, we have pseudo-labels $\hat{y}_{i}$ representing the cluster indices for all the images.
These pseudo-labels provide additional prior knowledge about the dataset, which help with formulating more reasonable positive pairs and negative pairs.

\subsection{Cluster-guided Contrastive Learning}
\label{ssec:ccl}
Here, we aim to train a model to learn image representations that take advantage of the clustering information above. 
Given a batch of input images $\left\{x_{i} | i=1, 2, \ldots, b\right\}$ and their pseudo-labels $\left\{\hat{y}_{i} | i=1, 2, \ldots, b\right\}$, we apply two different augmentations $\hat{x}_{i} = aug_{1}(x_{i})$ and $\tilde{x}_{i} = aug_{2}(x_{i})$, combining spatial and appearance transformations, to generate two different views. We then use an encoder to map the images to one-dimensional features and deploy an additional multi-layer perceptron (MLP) to increase the representation ability, $f_{i} = MLP(Enc(\hat{x_{i}}))$. We also employ $L_2$ normalization on the features $f_{i}$ for training stability. So far, the above processes are the same as SimCLR \cite{chen2020simple}, but the biggest difference in our method is our selection strategy for positive pairs. Since we obtain a pseudo-label for each image, we extend the concept of positive pairs to features that correspond to images categorized as in the same cluster. 
The contrastive loss for a batch is defined as follows:
\begin{equation}
\label{con_loss}
\small{
    L_{con} = -\frac{1}{2b}\sum_{ f_{i} \in B}\frac{1}{\left | P(f_{i}) \right |}\log\frac{\sum_{f_{j} \in P(f_{i})}\exp(f_{i}\cdot f_{j} / \tau )}{\sum_{f_{k} \in B - P(f_{i})}\exp(f_{i}\cdot f_{k} / \tau )}}
\end{equation}
where $B$ is the feature set of size $2*b$ combining the two augmentation views' features, $\tau$ is a temperature parameter, and $P(f_{i}) = \left\{{f_{j} | \hat{y}_{j}} = \hat{y}_{i}\right\}$ is the collection of features that are viewed as positive pairs of $f_{i}$. 
In comparison, $P(f_{i})$ in SimCLR \cite{chen2020simple} and MoCo \cite{chen2020mocov2} contains only a feature from the same image, which might result in many harmful negative pairs (e.g., images belonging to the same cluster). 
We also notice that BYOL \cite{grill2020bootstrap} and SimSiam \cite{chen2021exploring} inherently avoid the issue of harmful negatives by only utilizing positive pairs.
However, we elect to use a contrastive loss over these approaches because 1) they were designed for large-scale natural scene datasets while the number of images or slices in a medical image segmentation dataset is often much smaller, and 2) negative samples benefit medical images more due to more limited appearance variations \& semantic categories.

\section{Experiments and Results}
\label{sec:experiment}

\subsection{Datasets and Implementation Details}

\noindent
\textbf{Skin Lesion Segmentation}. 
For ISIC'2017 \cite{Gutman2018SkinLA}, we use the provided 2000 train, 150 validation, and 600 test image splits with a single foreground class. 
During pretraining, we additionally use training images from ISIC's 2018 (10015 images) and 2020 (33126 images) challenges for a total of 45141 unlabeled samples after removing duplicate images.
For preprocessing and augmentations, we normalize pixel intensities to have zero mean and unit variance, apply a random resized crop with a network input size of 192$\times$192, and follow additional augmentations used by BYOL \cite{grill2020bootstrap}.


\noindent
\textbf{Whole Heart Segmentation}. 
MMWHS'2017 \cite{Zhuang2013ChallengesAM} 
contains 20 CT volumes with 7 foreground classes.
We perform 4-fold cross-validation on the 20 volumes with a 2:1:1 train:validation:test ratio.
For preprocessing, we first crop the 3D images to keep only the areas of interest, and then we resize the shape of each slice along the axial plane to 256$\times$256 along with intensity normalization to values between 0 and 1. 

\noindent
\textbf{Implementation}. 
We run all experiments (coded in PyTorch) on NVIDIA P100s. 
For our backbone network, we use a vanilla UNet 
.
Only the down-sampling path of the architecture is pretrained, while we fine-tune using the transferred encoder weights with a randomly initialized decoder.
For the two-layer MLP head, we use 2048 intermediate and 256 output dimensions.
For network training on ISIC (resp., MMWHS) data, we employ AdamW 
with $\beta_{1} = 0.9$, $\beta_{2}=0.999$ for feature extraction, 
contrastive learning, and fine-tuning tasks with learning rates of $1\times10^{-3}$ (resp., $3\times10^{-4}$), $2.5\times10^{-4}$ ($3\times10^{-4}$), and $1\times10^{-4}$ ($5\times10^{-4}$), while training for 200 (200), 400 (200), and 200 (120) epochs, with batch sizes of 60 (48), 55 (48), and 24 (8), respectively.
We use cosine learning rate decay for both datasets and $\tau$=0.1.

\noindent


\subsection{Comparison with State-of-the-arts}

For each experiment in Tab. \ref{table1}, we pretrain exclusively on images in the training split and evaluate on the test split.
When evaluating the effectiveness of our proposed method versus state-of-the-art contrastive methods, we note the following observations.
%
%
%
%
%
\textbf{1)} Our framework \emph{beats recent natural scene contrastive methods} and improves the quality of representations for both datasets. 
This supports our hypothesis that reducing harmful positives \& negatives through appearance priors would be beneficial over vanilla instance discrimination.
\textbf{2)} Our method \emph{outperforms contrastive methods for 3D medical imaging} even when spatial assumptions reasonably hold (e.g., on the MMWHS dataset).
Here, LGCL and PCL are adversely impacted by both harmful positives when contents from corresponding slices are poorly aligned and harmful negatives where slices present similar tissues (e.g., disparate regions of large organs; see right side of Fig.~\ref{fig:motivation}). 
\textbf{3)} \emph{Negative samples play an important role in metric learning}. When comparing with BYOL, a method that uses only positive samples, we can see that the performance struggles to even beat training from scratch. 
We reiterate our point that negative samples are key since medical images tend to have less appearance diversity, categories, and amount of overall samples.

\subsection{Study on Clustering Effects}

The quality of extracted features and their clustering is central to our approach.
For feature learning, we empirically observe that the type of auto-encoder architecture (e.g., AE, VAE, $\beta$-VAE) had minimal effects on downstream performance.
For clustering, we see more pronounced effects from both clustering type and granularity (see Tab. \ref{tab:abl1}).
Granularity modulates the difficulty of negative samples (raising k eventually reduces to instance discrimination) and the number of harmful negatives present.
To initialize $k$, selecting the elbow point on the curve of mean distance within each cluster seems to be a reasonable heuristic that outperforms SimCLR.

\section{Conclusions}

In this paper, we proposed a pretraining framework that uses appearance priors through unsupervised feature clustering to facilitate contrastive learning of representations for medical images. 
Our approach requires no additional metadata or stringent spatial priors, and still outperforms recent state-of-the-art contrastive methods on both 2D skin lesion segmentation \& 3D CT cardiac structure segmentation.

\begin{table}
\begin{center}
\caption{\label{tab:abl1}Study on the effects of clustering type and categorical granularity on 5$\%$ annotated ISIC skin segmentation. 
    DBSCAN uses the conventional minpoints=$4$.}
\vspace*{-4mm}
\resizebox{\columnwidth}{!}{
\begin{tblr}{
    colspec={c ||c c c | c c c | c c c },
    row{1-2} = {gray!50!black!15},
    }
\hline 
\multirow{2}{*} & \multicolumn{3}{c} {k-Means} & 
\multicolumn{3}{c} {DBSCAN} &   
\multicolumn{3}{c} {Agglomerative} \\
\cline{2-12} 
& $k$=3 & $k$=20 & $k$=100 & $\epsilon$=0.9 & $\epsilon$=0.5 & $\epsilon$=0.1 & $k$=3 & $k$=20 & $k$=100 \\
\hline  
Dice \% & 79.1 & 80.2 & 80.4 & 
                  78.5 & 79.6 & 80.0 & 
                  79.0 & \textbf{81.9} & 80.5  \\
\hline
\end{tblr}
}
\end{center}
\vspace{-6mm}
\end{table}



\bibliographystyle{plain}
\bibliography{refs}{}

\end{document}